# FACE DETECTION USING ADABOOSTED SVM-BASED COMPONENT CLASSIFIER

**Seyyed Majid Valiollahzadeh, Abolghasem Sayadiyan**
*Electrical Engineering Department, Amirkabir University of Technology,
Tehran, Iran, 15914*
valiollahzadeh@yahoo.com, eea35@aut.ac.ir

**Mohammad Nazari**
*Electrical Engineering Department, Amirkabir University of Technology,
Tehran, Iran, 15914*
mohnazari@aut.ac.ir

Abstract: Recently, Adaboost has been widely used to improve the accuracy of any given learning algorithm. In this paper we focus on designing an algorithm to employ combination of Adaboost with Support Vector Machine (SVM) as weak component classifiers to be used in Face Detection Task. To obtain a set of effective SVM-weaklearner Classifier, this algorithm adaptively adjusts the kernel parameter in SVM instead of using a fixed one. Proposed combination outperforms in generalization in comparison with SVM on imbalanced classification problem. The proposed here method is compared, in terms of classification accuracy, to other commonly used Adaboost methods, such as Decision Trees and Neural Networks, on CMU+MIT face database. Results indicate that the performance of the proposed method is overall superior to previous Adaboost approaches.

## 1 INTRODUCTION

Nonlinear classification of data is always of special attention. Face Detection is a problem dealing with such data, due to large amount of variation and complexity brought about by changes in facial appearance, lighting and expression. Feature selection is needed beside appropriate classifier design to solve this problem, like many other pattern recognition tasks.

One of the major developments in machine learning in the past decade is the Ensemble method, which finds a highly accurate classifier by combining many moderately accurate component classifiers. Two of the commonly used techniques for constructing Ensemble classifiers are Boosting [schaphire, 2002] and Bagging [Breiman, 1996]. In Comparison with Bagging, Boosting outperforms when the data do not have much noise [Opitz, 1999] [Bauer, 1999] .Among popular Boosting methods, AdaBoost [Freund, 1997] establishes a collection of weak component classifiers by maintaining a set of weights over training samples and adjusting them adaptively after each Boosting iteration: the weights of the misclassified samples by current component classifier will be increased while the weights of the correctly classified samples will be decreased. To implement the weight updates in Adaboost, several algorithms have been proposed [Kuncheva, 2002]. The success of AdaBoost can be attributed to its ability to enlarge the margin [schapire, 1998], which could enhance AdaBoost's generalization capability. Decision Trees [Dietterich, 2000] or Neural Networks [Schwenk, 2000] [Ratsch, 2001] have already been employed as component classifiers for AdaBoost. These studies showed good generalization performance of these AdaBoost. However, determining the suitable tree size is a question when Decision Trees are used as component classifiers. Also, controlling the complexity in order to avoid over fitting will remain a question, when Radial Basis Function (RBF) Neural Networks are used as component classifiers.

Moreover, we have to decide on the optimum number of centers and also on setting the width values of the RBFs. All these factors have to be carefully tuned in practical use of AdaBoost. Furthermore, diversity is known to be an important factor which affects the generalization accuracy of Ensemble classifiers [Melville, 2005][Kuncheva, 2002]. In order to quantify the diversity, some methods are proposed [Kuncheva, 2003] [Windeatt, 2005]. It is also known that in AdaBoost exists an accuracy/diversity dilemma [Dietterich, 2000], which means that the more accurate two component classifiers become, the less they can disagree with each other. Only when the accuracy and diversity are well balanced, can the AdaBoost demonstrate excellent generalization performance. However, the existing AdaBoost algorithms do not yet explicitly taken sufficient measurement to deal with this problem. Support Vector Machine [Vapnick, 1998] was developed based on the theory of Structural Risk Minimization. By using a kernel trick to map the training samples from an input space to a high dimensional feature space, SVM finds an optimal separating hyper plane in the feature space and uses a regularization parameter, C, to control its model complexity and training error. One of the popular kernels used by SVM is the RBF kernel, including a parameter known as Gaussian width, $\sigma$. In contrast to the RBF networks, SVM with the RBF kernel (RBFSVM in short) can automatically determine the number and location of the centers and the weight values [Scholkopf, 1997]. Also, it can effectively avoid over fitting by selecting proper values of C and $\sigma$. From the performance analysis of RBFSVM [Valentini, 2004], we know that $\sigma$ is a more important parameter compared to C: although RBFSVM cannot learn well when a very low value of C is used, its performance largely depends on the $\sigma$ value if a roughly suitable C is given. This means that, over a range of suitable C, the performance of RBFSVM can be conveniently changed by simply adjusting the value of $\sigma$.

The proposed here method is compared, in terms of classification accuracy, to other commonly used Adaboost methods, such as Decision Trees and Neural Networks, on CMU+MIT face database. Results indicate that the performance of the proposed method is overall superior to those of traditional adaboost approaches.

## 2 FEATURE SELECTION

In this paper, like Viola and Jones [Viola and Jones 2001], we use four types of Haar-like basis functions for feature selection which have been used by Papageorgiou et al [Papageorgiou et al 1998].
Like their work, we use four types of haar-like feature to build the feature pool. The feature can be computed efficiently with integral image. The main objective to use these features is that they can be rescaled easily which avoids to calculate a pyramid of images and yields to fast operation of the system on these features. These features can be seen in figure1. Given that the base resolution of the detector is 32x32, the exhaustive set of rectangle features is quite large, over 180,000. Note that unlike the Haar basis, the set of rectangle features is overcomplete. For each scale level, we rescale the features and record the relative coordinate of the rescaled features to the top-left of integral image in look-up-table (LUT). After looking up the value of the rescaled rectangle's coordinate, we calculate features with relative coordinate. Like viola, we use image variance $\sigma$ to correct lighting, which can be got using integral images of both original image and image squared. Rescaling needs to round rescaled coordinates to nearest integer, which would degrade the performance of viola's features [Lienhart 2003]. Like R. Lienhart [Lienhart 2003], we normalize the features by acreage, and thus reduce the rounding error.

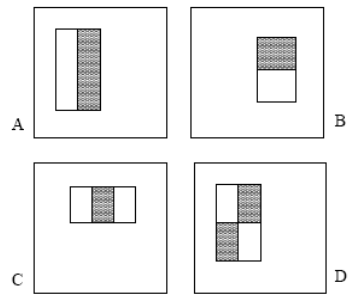

Figure 1: Example rectangle features shown relative to the enclosing detection window. The sum of the pixels which lie within the white rectangles is subtracted from the sum of pixels in the grey rectangles. Two-rectangle features are shown in (A) and (B). Figure (C) shows a three-rectangle feature, and (D) a four-rectangle feature.

Using the integral image any rectangular sum can be computed in four array references (see Figure 2). Clearly the difference between two rectangular sums can be computed in eight references. Since the two-rectangle features defined above involve adjacent rectangular sums they can be computed in six array references, eight in the case of the three-rectangle features, and nine for four-rectangle features.

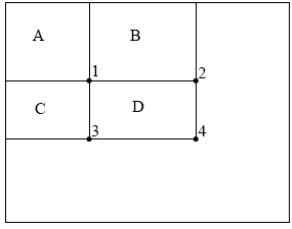

Figure 2: The sum of the pixels within rectangle D can be computed with four array references. The value of the integral image at location 1 is the sum of the pixels in rectangle A. The value at location 2 is A+B, at location 3 is A+C, and at location 4 is A+B+C+D. The sum within D can be computed as 4+1-(2+3).

## 3 STATISTICAL LEARNING

In this section, we describe boost based learning methods to construct face/nonface classifier, and propose a new boosting algorithm which improves boosting learning.

### 3.1 AdaBoost Learning

Given a set of training samples, AdaBoost [Schapire and Singer 1999] maintains a probability distribution, W, over these samples. This distribution is initially uniform. Then, AdaBoost algorithm calls Weak Learn algorithm repeatedly in a series of cycles. At cycle T, AdaBoost provides training samples with a distribution $w^t$ to the WeakLearn algorithm.

AdaBoost, constructs a composite classifier by sequentially training classifiers while putting more and more emphasis on certain patterns.

For two class problems, we are given a set of N labeled training examples $(y_1, x_1),...,(y_N, x_N)$, where $y_i \in \{+1,-1\}$ is the class label associated with example $x_i$.

For face detection, $x_i$ is an image sub-window of a fixed size (for our system 24x24) containing an instance of the face $(y_i = +1)$ or non-face $(y_i = -1)$ pattern. In the notion of AdaBoost see Algorithm 1, a stronger classifier is a linear combination of M weak classifiers.

In boosting learning [9, 26, 10], each example $x_i$ is associated with a weight $w_i$, and the weights are updated dynamically using a multiplicative rule according to the errors in previous learning so that more emphasis is placed on those examples which are erroneously classified by the weak classifiers learned previously.

Greater weights are given to weak learners with lower errors. The important theoretical property of AdaBoost is that if the weak learners consistently have accuracy only slightly better than half, then the error of the final hypothesis drops to zero exponentially fast. This means that the weak learners need be only slightly better than random.

Furthermore, since proposed AdaBoost with SVM invents a convenient way to control the classification accuracy of each weak learner, it also provides an opportunity to deal with the well-known accuracy/diversity dilemma in Boosting methods. This is a happy accident from the investigation of AdaBoost based on SVM weak learners.

---

Algorithm 1. The AdaBoosAlgorithm [Schapire and Singer] .

1. Input: Training sample
Input: a set of training samples with labels $(y_1, x_1),...,(y_N, x_N)$, ComponentLearn algorithm, the number of cycles $T$.

2. Initialize: the weights of training samples: $w_i^1 = 1/N$, for all $i = 1,...,N$

3. Do for $t = 1,...,T$
   (1) Use ComponentLearn algorithm to train the component classifier $h_t$ on the weighted training sample set.
   (2) Calculate the training error of $h_t$:
   $$\varepsilon_t = \sum_{i=1}^{N} w_i^t, y_i \neq h_t(x_i).$$
   (3) Set weight of component classifier $h_t$:
   $$h_t : \alpha_t = \frac{1}{2}\ln\left(\frac{1-\varepsilon_t}{\varepsilon_t}\right)$$
   (4) Update the weights of training samples:
   $$w_i^{t+1} = \frac{w_i^t \exp\{-\alpha_t y_i h_t(x_i)\}}{C_t}$$
   where $C_t$ is a normalization constant, and
   $$\sum_{i=1}^{N} w_i^{t+1} = 1$$

4. Output: $f(x) = sign(\sum_{t=1}^{T} \alpha_t h_t(x))$.

---

### 3.2 SVM Based Approach for Classification

The principle of Support Vector Machine (SVM) relies on a linear separation in a high dimension feature space where the data have been previously mapped, in order to take into account the eventual non-linearities of the problem.

If we assume that, the training set $X = (x_i)_{i=1}^{l} \subset R^R$ where $l$ is the number of training vectors, R stands for the real line and R is the number of modalities, is labelled with two class targets $Y = (y_i)_{i=1}^{l}$, where :

$$y_i \in \{-1, +1\} \quad \Phi : R^R \to F \qquad (1)$$

Maps the data into a feature space F. Vapnik has proved that maximizing the minimum distance in space F between $\Phi(X)$ and the separating hyper plane $H(w,b)$ is a good means of reducing the generalization risk.

Where:

$$H(w,b) = \{f \in F \mid <w, f>_F + b = 0\}, \qquad (2)$$
$$(<> \text{ is inner product})$$

Vapnik also proved that the optimal hyper plane can be obtained solving the convex quadratic programming (QP) problem:

$$\text{Minimize} \quad \frac{1}{2}\|w\|^2 + c\sum_{i=1}^{l}\xi \qquad (3)$$
$$\text{with} \quad y_i(<w, \Phi(X)> + b) \geq 1 - \xi \quad i = 1, ..., l$$

Where constant C and slack variables x are introduced to take into account the eventual non-separability of $\Phi(X)$ into F.

In practice this criterion is softened to the minimization of a cost factor involving both the complexity of the classifier and the degree to which marginal points are misclassified, and the tradeoff between these factors is managed through a margin of error parameter (usually designated C) which is tuned through cross-validation procedures.

Although the SVM is based upon a linear discriminator, it is not restricted to making linear hypotheses. Non-linear decisions are made possible by a non-linear mapping of the data to a higher dimensional space. The phenomenon is analogous to folding a flat sheet of paper into any three-dimensional shape and then cutting it into two halves, the resultant non-linear boundary in the two-dimensional space is revealed by unfolding the pieces.

The SVM's non-parametric mathematical formulation allows these transformations to be applied efficiently and implicitly: the SVM's objective is a function of the dot product between pairs of vectors; the substitution of the original dot products with those computed in another space eliminates the need to transform the original data points explicitly to the higher space. The computation of dot products between vectors without explicitly mapping to another space is performed by a kernel function.

The nonlinear projection of the data is performed by this kernel functions. There are several common kernel functions that are used such as the linear, polynomial kernel $(K(x,y) = (<x, y>_{R^R} + 1)^d$ and the sigmoidal kernel $(K(x,y) = \tanh(<x, y>_{R^R} + a))$, where x and y are feature vectors in the input space.

The other popular kernel is the Gaussian (or "radial basis function") kernel, defined as:

$$K(x,y) = \exp(\frac{-|x-y|^2}{(2\sigma^2)}) \qquad (4)$$

Where $\sigma$ is a scale parameter, and x and y are feature-vectors in the input space. The Gaussian kernel has two hyper parameters to control performance C and the scale parameter $\sigma$. In this paper we used radial basis function (RBF).

### 3.3 AdaBoosted SVM-Based Component Classifier

We combine SVM with AdaBoost to improve its capability in classification. When applying Boosting method to strong component classifiers, these classifiers must be appropriately weakened in order to benefit from Boosting [Dietterich 2000].

Like Schapire and Singer, we used resampling to train AdaBoost, in this problem we must train weak classifiers (SVM classifier) to obtain best Gaussian width, σ and the regularization parameter, C, for optimizing strong classifier (AdaBoost classifier).

Hence, SVM with RBF kernel is used as weak learner for AdaBoost, a relatively large σ value, which corresponds to a SVM with RBF kernel with relatively weak learning ability, is preferred. Both resampling and reweighting can be used to train AdaBoost. The algorithm is shown in the following diagram.

Algorithm 2. The AdaBoost with SVM Algorithm.

1. Input: Training sample
Input: a set of training samples with labels $(y_1, x_1), ..., (y_N, x_N)$,

The initial $\sigma = \sigma_{ini}$, $\sigma_{min}$, $\sigma_{step}$

2. Initialize: the weights of training samples: $w_i^1 = 1/N$, for all $i = 1, ..., N$

3. Do while $\sigma > \sigma_{min}$
   (1) Use RBFSVM to train on the weighted training sample set.
   (2) Calculate the training error of $h_t$:
   $$\varepsilon_t = \sum_{i=1}^{N} w_i^t, y_i \neq h_t(x_i).$$
   (3) if $\varepsilon_t > .5$, decrease $\sigma$ value by $\sigma_{step}$ and goto(1)
   (4) Set weight of component classifier $h_t$:
   $$h_t : \alpha_t = \frac{1}{2} \ln\left(\frac{1-\varepsilon_t}{\varepsilon_t}\right)$$
   (5) Update the weights of training samples:
   $$w_i^{t+1} = \frac{w_i^t \exp\{-\alpha_t y_i h_t(x_i)\}}{C_t}$$
   where $C_t$ is a normalization constant, and
   $$\sum_{i=1}^{N} w_i^{t+1} = 1$$

4. Output: $f(x) = sign(\sum_{t=1}^{T} \alpha_t h_t(x)).$

## 4 EXPERIMENTAL RESUTTS

### 4.1 Database

We tested our system on the MIT+CMU frontal face test set [Rowley et al. 1994] and own database. There are more than 2,500 faces in total. To train the detector, a set of face and nonface training images were used. The pairwise recognition framework is evaluated on a compound face database with 2000 face images hand labelled faces scaled and aligned to a base resolution 32 by 32 pixels by the centre point of the two eyes and the horizontal distance between the two eyes. For non-face training set, an initial 10,000 non-face samples were selected randomly from 15,000 large images which contain no face.

### 4.2 Face Detection System

We explain our face detection system and show how to construct a AdaBoosted SVM-based component classifier for face detection. The learning of a detector is done as follows:

1. A set of simple Haar wavelet features are used as candidate features. There are tens of thousands of such features for a 32x32 window.
2. A subset of them are selected and the corresponding weak classifiers are constructed, using AdaBoosted SVM-based component classifier learning.
3. A strong classifier is constructed as a linear combination of the weak ones.
4. A detector is composed of one or several strong classifiers in cascade.

The detector pyramid is then built upon the learned detectors [Li and Zhang 2004].

### 4.3 Results

The SVM-based component classifier and AdaBoost algorithm are used for the classification of each pair of individuals. We compare the detection rates to other commonly used Adaboost methods, such as Decision Trees and Neural Networks, on face database.

For showing the performance of our AdaBoosted svm-based component classifier algorithm, the results are shown in Table 1.

| False detections / Detector | 120 | 200 |
|---|---|---|
| Adaboost with SVM | 5.41 | 1.85 |
| Adaboost with Decision Trees | 9.81 | 2.42 |
| Adaboost with Neural Networks | 14.51 | 5.41 |

Table 1: Comparison of Error rate (%) for some AdaBoost methods.

A ROC curve showing the performance of our detector on this test set is shown in Figure 3 and Some results are shown in Figure 4.

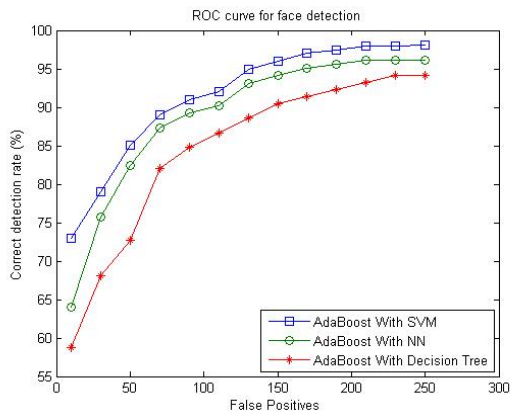

Figure 3: Comparison of ROC for frontal face detection results.

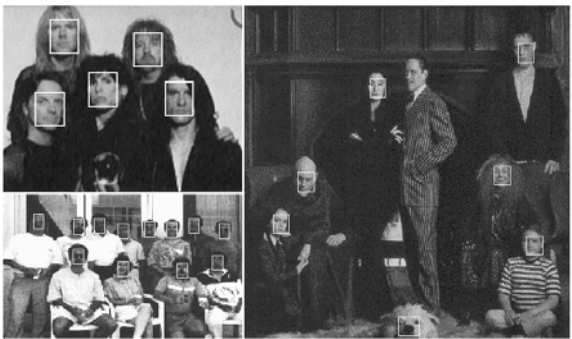

Figure 4: Some frontal face detection results.

## 5 CONCLUSIONS

AdaBoost with properly designed SVM-based component classifiers is proposed in this paper, which is achieved by adaptively adjusting the kernel parameter to get a set of effective component classifiers. Experimental results on CMU+MIT database for Face Detection demonstrated that proposed AdaBoostSVM algorithm performs better than other approaches of using component classifiers such as Decision Trees and Neural Networks in accuracy and speed. Besides these, it is found that proposed AdaBoostSVM algorithm demonstrated good performance on imbalanced classification problems. Based on the AdaBoostSVM, an improved version is further developed to deal with the accuracy/diversity dilemma in Boosting algorithms, giving rising to better generalization performance. Experimental results indicate that the performance of the cascaded adaboost classifier with SVM is overall superior to those obtained by the NN and Decision Tree.

## ACKNOWLEDGEMENTS

The authors would like to acknowledge the Iran Telecommunication Research Center (ITRC) for financially supporting this work.